\documentclass[journal]{IEEEtran}
\ifCLASSINFOpdf
\else
\fi
\usepackage[hidelinks]{hyperref}   
\usepackage{hyperref}
\usepackage{subcaption}
\usepackage{epstopdf}
\usepackage{graphicx}
\usepackage{amsfonts}   
\usepackage{enumerate}
\usepackage{varioref}
\usepackage{longtable}
\usepackage{graphicx}
\usepackage{amssymb}
\usepackage{chngpage}
\usepackage{multirow}
\usepackage{multirow}
\usepackage{amsthm}
\usepackage{amsmath}
\usepackage{caption}
\usepackage{nccmath}  
\usepackage[table]{xcolor}  
\usepackage{algorithm}
\usepackage{algpseudocode}
\usepackage{placeins}  
\usepackage{url}
\usepackage{threeparttable}

\begin{document}

\title{Pixel-Level Alignment of Facial Images\\ for High Accuracy Recognition\\ Using Ensemble of Patches}

\author{Hoda Mohammadzade,
		Amirhossein Sayyafan,
        Benyamin Ghojogh

\thanks{Hoda Mohammadzade's e-mail: hoda@sharif.edu}        
\thanks{Amirhossein Sayyafan's e-mail: amirhossein.sayyafan@alum.sharif.edu}
\thanks{Benyamin Ghojogh's e-mail: \href{mailto:ghojogh_benyamin@ee.sharif.edu}{ghojogh\_benyamin@ee.sharif.edu}}
\thanks{All authors are with Department of Electrical Engineering, Sharif University of Technology, Tehran, Iran.}}


\maketitle

\begin{abstract}
The variation of pose, illumination and expression makes face recognition still a challenging problem. As a pre-processing in holistic approaches, faces are usually aligned by eyes. The proposed method tries to perform a pixel alignment rather than eye-alignment by mapping the geometry of faces to a reference face while keeping their own textures. The proposed geometry alignment not only creates a meaningful correspondence among every pixel of all faces, but also removes expression and pose variations effectively. The geometry alignment is performed pixel-wise, i.e., every pixel of the face is corresponded to a pixel of the reference face. 
In the proposed method, the information of intensity and geometry of faces are separated properly, trained by separate classifiers, and finally fused together to recognize human faces. Experimental results show a great improvement using the proposed method in comparison to eye-aligned recognition. For instance, at the false acceptance rate of 0.001, the recognition rates are respectively improved by 24\% and 33\% in Yale and AT\&T datasets. In LFW dataset, which is a challenging big dataset, improvement is 20\% at FAR of 0.1. 
\end{abstract}


\IEEEpeerreviewmaketitle
\section{Introduction}\label{Introduction_section}

Face Recognition is one of the most attractive and practical fields of research in pattern analysis and image processing, receiving much attention from different knowledge backgrounds including pattern recognition, computer vision, image processing, statistical learning, neural networks, and computer graphics \cite{zhao2003face}.

According to \cite{zhao2003face}, face recognition methods can be categorized into two main categories; feature-based and holistic (whole-pixels) methods. Feature-based methods try to create a feature vector out of the face for the learning process. The holistic recognition uses all pixels of face region as raw data for recognition and learning.

Feature-based methods utilize the geometrical and structural features of face \cite{zhao2003face}. For instance, in \cite{kelly1970visual}, features of head width, distances between eyes and eyes to mouth are compared. 
In \cite{kanade1977computer}, angles and distances between eye corners, mouth hole, chin top and the nostrils are used. 
In \cite{brunelli1993face}, face features such as mouth, nose, eyebrows, and face outline are detected using horizontal and vertical gradients. In this method, template matching using correlation is also proposed.
In \cite{nefian1998hidden,samaria1994hmm} Hidden Markov Model (HMM) is used on pixel strips of different parts of face. 
Also, recently, a patch-based representation is used in \cite{ding2015multi} in which each patch tries to learn a transformation dictionary in order to transform the features onto a discriminative subspace. 
Paper \cite{chen2013blessing} is another feature-based method in which a pyramid of facial image is created and the patches around five key landmarks in different pyramid levels are concatenated to prepare a high-dimensional feature vector.

Some feature-based methods use both features and whole pixels together in order to enhance the performance of recognition \cite{zhao2003face}. Eigenmodules \cite{pentland1994view} can be mentioned in which eigenfaces are combined with eigenmodules of face such as eigeneyes, eigenmouth and eigennose. In \cite{penev1996local} Principle Component Analysis (PCA) is used in combination with Local Feature Analysis (LFA). 
Some of the methods in this category which seem more promising are based on \textquotedblleft shape-free\textquotedblright\text{} face concept.
In \cite{cootes1998active,cootes2001active}, Active Appearance Model (AAM) has been proposed as a method of warping textures of image patches to a specific geometry in an iterative manner. In this method \cite{lanitis1995automatic,stegmann2002analysis,edwards1998face,lanitis1995unified}, the patch of the face is labeled by several landmarks (model points), the texture of face is projected onto the texture model frame by applying scale and offset to the intensities, and the residual (error) between the projected and previous image patches is iteratively reduced.
In \cite{lanitis1995automatic,lanitis1995unified}, the shape of the face is also modeled using Active Shape Model (ASM) \cite{cootes1995active}. The authors have shown that different weights of eigenvalues can vary the different aspects and parts of face shape models.
In \cite{craw1992face}, several shape-free (neutral) faces create an ensemble and all the faces are approximated by a linear combination of the eigenfaces of the ensemble.

Despite significant advances of feature-based methods, holistic methods are still being received lots of attention as they use the information of all pixels in the face region. Holistic methods detect and crop the face out of the image and use it as a raw input for classification. Eigenfaces \cite{turk1991eigenfaces,turk1991face}, Fisherfaces \cite{belhumeur1997eigenfaces}, and Kernel faces \cite{lu2003face,lu2005kernel} are several well-known examples of this category which respectively create a feature space using Principle Component Analysis (PCA), Fisher Linear Discriminant Analysis (LDA) and Kernel Direct Discriminant Analysis (KDDA) for face classification and recognition.
Face recognition using support vector machine (SVM) \cite{phillips1999support} is another method from this category, which formulates face recognition as a two-class problem, one class as dissimilarities between faces of the same person and the other class as dissimilarities between faces of different individuals.
Bayesian classifier \cite{moghaddam2000bayesian} can also be mentioned in this category, which has a probabilistic approach toward the similarity of faces.
Some other holistic methods of face recognition have used artificial neural networks \cite{zhao2003face,kasar2016face}. As instance, Probabilistic Decision-Based Neural Network (PDBNN) \cite{lin1997face} and Convolutional Neural Networks (CNN) \cite{simon2016improved,abdalmageed2016face,parkhi2015deep,taigman2014deepface,schroff2015facenet} can be mentioned. 
Recently, Sparse Representation based Classification (SRC) \cite{wright2009robust} is used in order to create a recognition system with robustness to illumination and occlusion. 

Both geometrical and intensity features exist in a 2D image of a face, which help human eye to recognize people from their images. E.g., both eye color and the distance between eyes and nose can be inferred from a facial image. Accordingly, in a successful face recognition system both of these categories of features, i.e., geometry and intensity, should be appropriately used. However, whenever eye alignment is used in a holistic method or other approaches, the correspondence between organs other than eyes are disturbed; the intensity of lips in different faces cannot be compared with each other nor their position can be compared. The main contribution of this work is to introduce an alignment method by which the intensity and geometry information are separated from each other. Each of these pieces of information is then used to train their corresponding classification modules and finally their results are fused together to recognize human faces. Moreover, note that any classification algorithm, such as Fisher-LDA-based ensemble of patches which is used in this paper, can be used as the classifier in the proposed method. In particular, the proposed method provides appropriate inputs for methods such as Convolutional Neural Networks (CNN), which automatically design specialized filters and require aligned raw features as inputs.

To mention in more details, two major tasks are accomplished using the proposed method:
\begin{enumerate}
\item The proposed alignment method places the intensity of similar organs in the same positions in the warped faces.
\item When intensities are properly aligned, using the proposed geometry extraction method, the coordinate of the aligned pixels can be extracted to be used as the corresponding geometry information. 
\end{enumerate}
As a result, the proposed pixel alignment provides both intensity and geometry information useful for recognition.

The remainder of this paper is organized as follows. Section \ref{Geometrical_Alignment_section} details the geometrical alignment as the first part of proposed method. Thereafter, geometrical information is more discussed in Section \ref{Geometrical_Information_section}.
Creating feature vectors using ensemble of patches, using Fisher Linear Discriminant Analysis (LDA), and decision fusion are explained in Section \ref{Classification_section} afterwards. 
Section \ref{Classification_section} also sums up the proposed method by illustrating the overall structure.
The utilized datasets and experimental results are also reported in Section \ref{Experimental_results_section}. 
Some discussions on alignment of features and ensemble of patches are gone through in Section \ref{Discussions_section}.
Finally, in Sections \ref{Conclusion_section} and \ref{Future_work_section}, article is concluded and the potential future work is mentioned, respectively.

\section{Geometrical Alignment}\label{Geometrical_Alignment_section}
 
Geometrical alignment can be defined as aligning the geometries of faces to a unique geometry while saving their own textures.
In the proposed geometrical alignment method, a reference geometry is defined and the geometry of all faces of train and test procedures are transformed to this geometry. 
Here, the geometry of a face is defined as the location of the contours of the facial landmarks. Therefore, geometrical alignment is performed by warping a face such that its facial contours coincide with those of the reference contours.

In the proposed method, in order to detect facial landmarks, every landmark detection method can be used such as Active Shape Model (ASM) \cite{cootes1995active} or Constrained Local Neural Fields (CLNF) \cite{baltrusaitis2013constrained}. In this work, CLNF is utilized for this purpose.\footnote{The code of CLNF method can be found in https://github.com/TadasBaltrusaitis/OpenFace.}
The landmarks in this work are as follows. There are 17 landmarks around the face region, 14 landmarks for lips, three landmarks for each upper and lower teeth, six landmarks for each eye, nine landmarks for the whole nose and five landmarks for each eyebrow, resulting in 68 total landmarks.

In the following sections, different steps of the proposed method are explained in details.

\subsection{Fitting Face contours}


The CLNF method \cite{baltrusaitis2013constrained}, which is an enhanced Constrained Local Model (CLM) \cite{cristinacce2006feature}, is used for detecting landmarks of each train and test face. 
This method is briefly described in the following. Interested readers are encouraged to refer to \cite{baltrusaitis2013constrained} for more details.

The CLNF method consists of two main parts: (I) probabilistic patch expert (landmark detector), and (II) non-uniform regularized landmark mean-shift optimization technique.

At first, face or faces are detected with a tree-based method. CLNF method introduces patch experts which are small partitions of pixels around the interest points such as face edge, eyes, eyebrows, nose and lips.
The initial patch experts are put on the image. The pixels which fall in the $m^{th}$ patch are named as $X_m = \{x_1, x_2, ..., x_n\}$ where $x_i$ is a two-dimensional vector representing the coordinate of pixels.
This method uses a one-layer neural network with $X_m$ as inputs, and outputs $Y_m = \{y_1, y_2, ..., y_n\}$ where $y_i$ is a scalar \cite{baltrusaitis2013constrained}. A potential function is defined as $\Psi$ which is a function of vertex features $f_k$ and edge features $g_k$ and $l_k$. The features are defined as \cite{baltrusaitis2013constrained},

\begin{equation}
f_k(y_i, x_i, \theta_k) = -(y_i - h(\theta_k, x_i))^2,
\end{equation}

\begin{equation}
g_k(y_i, y_j) = -0.5 S_{ij}^{(g_k)} (y_i - y_j)^2,
\end{equation}

\begin{equation}
l_k(y_i, y_j) = -0.5 S_{ij}^{(l_k)} (y_i + y_j)^2,
\end{equation}
where $\theta_k$'s are the weights of $k^{th}$ neuron and $h(.,.)$ is the sigmoid activation function of the neural network. $S^{(g_k)}$ and $S^{(l_k)}$ control the smoothness (similarity) and sparsity, respectively.
This method attempts to maximize the probability

\begin{equation}
P(y|X) = \frac{e^\Psi}{\int_{-\infty}^{\infty} e^\Psi dy}.
\end{equation}

Non-uniform regularized landmark mean-shift optimization technique considers that variance of different patches are not similar and therefore sets several weights $W$'s for them. The contour $p$ of patches is updated as \cite{baltrusaitis2013constrained},

\begin{equation}
p^{new} = p^{old} + \Delta p,
\end{equation}
in which the step $\Delta p$ is defined as,

\begin{equation}
\Delta p = -(J^T W J + r \Lambda^{-1})(r \Lambda^{-1} p - J^T W v),
\end{equation}
where $J$ is the Jacobian of the landmark locations, $r$ is the regularization factor, $\Lambda^{-1}$ is the matrix describing the prior on the parameter $p$, and $v$ is the mean-shift vector over the patch responses \cite{baltrusaitis2013constrained}.

\subsection{Reference Contour}

\begin{figure}[!t]
\centering
\includegraphics[width=3in]{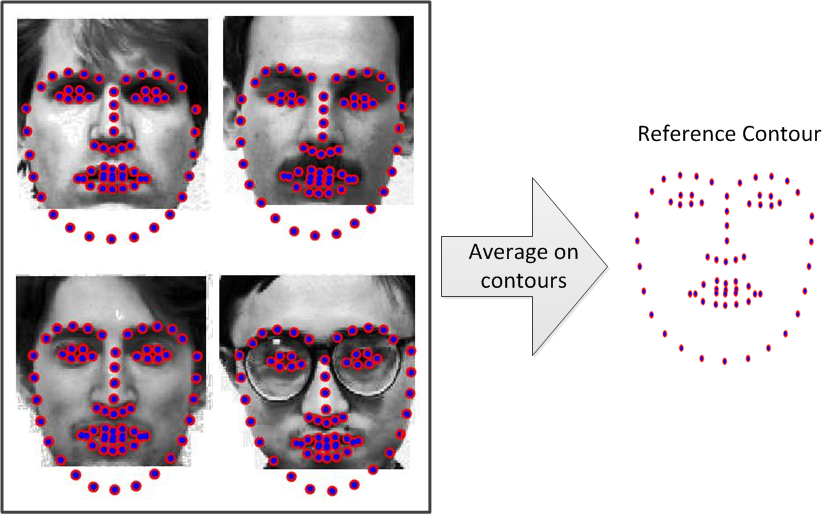}
\caption{Obtaining reference contour by averaging landmark contours of several neutral faces.}
\label{reference_face_contour}
\end{figure}

Reference contours are obtained by averaging the contours of landmarks of several neutral faces from the training set. Figure \ref{reference_face_contour} shows an example of reference contours.

\begin{figure}[!t]
\centering
\includegraphics[width=3in]{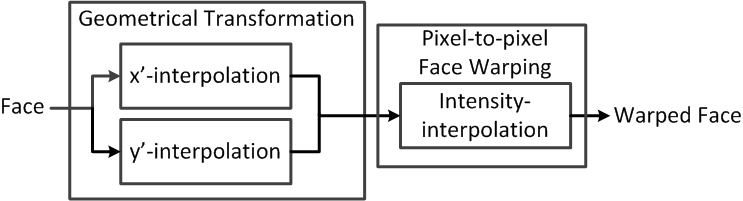}
\caption{Procedure of geometrical transformation and pixel-to-pixel face warping.}
\label{warp}
\end{figure}

\subsection{Transformation and Pixel-to-pixel Warping}
After fitting the contour of landmarks to the input face, the face is geometrically transformed and warped to reshape to the geometry of reference face. This step is detailed in this section.

For the geometrical transformation and pixel-to-pixel warping, three interpolations are performed as depicted in Fig. \ref{warp} which are detailed next. 
As a result of these interpolations, the intensity of each pixel is transformed to its corresponding location on the warped face. This transformation is guided by the transformation between the location of landmarks on the input face and the location of those on the warped face.

It is important to note that the proposed face warping method differs from the conventional one as the target coordinates for every single pixel of the input face is calculated using the described interpolation procedures.

\subsubsection{Affine Interpolation}

\begin{figure}[!t]
\centering
\includegraphics[width=2in]{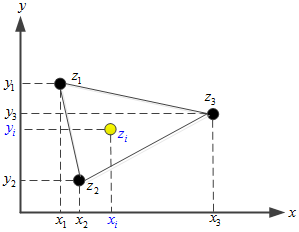}
\caption{Affine interpolation.}
\label{affine}
\end{figure}

In this work, affine transformation is used in order to perform the coordinate interpolations, i.e., $x'$- and $y'$-interpolation. Affine transformation uses three surrounding points to calculate the interpolated value at a new point. Assuming that the points have two dimensions. In affine interpolation method \cite{prince2012computer}, the value at each point is approximated as,

\begin{equation}
z_i = f(x_i,y_i) \simeq a_0 + a_1 x_i + a_2 y_i,
\end{equation}
where the coefficients $a_0$ to $a_2$ are calculated by solving the following linear system, according to Fig. \ref{affine},

\begin{equation}\label{eq_interpolate}
\begin{bmatrix}
    1 & x_1 & y_1 \\
    1 & x_2 & y_2 \\
    1 & x_3 & y_3 
\end{bmatrix}
\begin{bmatrix}
    a_0 \\
    a_1 \\
    a_2
\end{bmatrix}
=
\begin{bmatrix}
    z_1 \\
    z_2 \\
    z_3 
\end{bmatrix}.
\end{equation}
If this matrix equation is denoted as $XA=Z$, then by solving it using least square method, the coefficients are found as,

\begin{equation}
A=(X^T X)^{-1}X^T Z.
\end{equation}

\subsubsection{Delaunay Triangulation of Landmarks}

\begin{figure}[!t]
\centering
\includegraphics[width=1.75in]{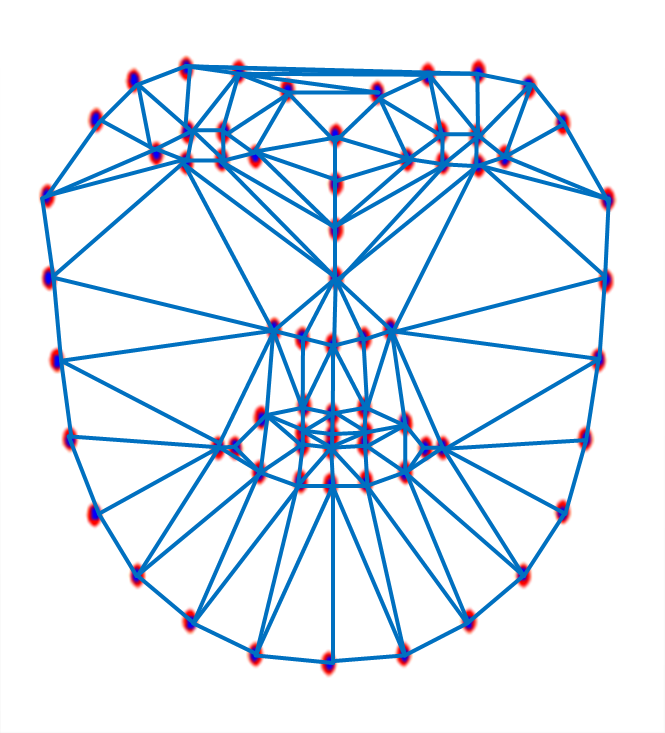}
\caption{Delaunay triangulation of face landmarks.}
\label{Delaunay}
\end{figure}

According to \cite{gartner2014computational}, a triangulation of a finite point set $P \subset \mathbb{R}^2$ is called a Delaunay triangulation, if the circumcircle of every triangle is empty, that is, there is no point from $P$ inside the circumcircle of any triangle.

Each face is triangulated using Delaunay method, as depicted in Fig. \ref{Delaunay}. By performing triangulation, the triangles needed for affine interpolations are obtained which are used in geometrical transformation as described next.

\subsubsection{Geometrical Transformation}

Let $(x,y)$ and $(x',y')$ denote the coordinates of pixels on the input and warped face, respectively, and $I(x,y)$ and $I'(x',y')$ denote their corresponding intensities.

\begin{figure}[!t]
\centering
\includegraphics[width=3.45in]{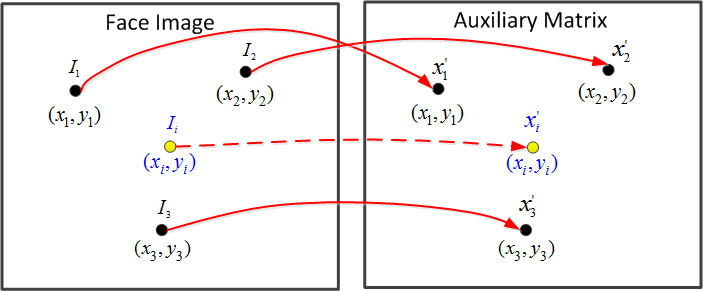}
\caption{$x'$-interpolation.}
\label{x_interpolation}
\end{figure}

\begin{figure}[!t]
\centering
\includegraphics[width=3.45in]{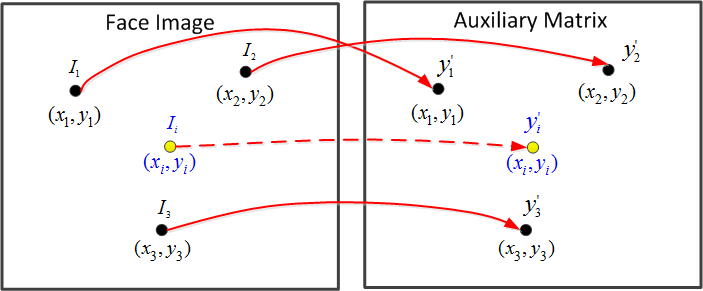}
\caption{$y'$-interpolation.}
\label{y_interpolation}
\end{figure}

For $x'$-interpolation, an auxiliary matrix is created with the same size as the input face. In this matrix, the $x'$ of landmarks are put on the same entry as they were in the input face matrix.
The other entries of this matrix are found using affine interpolation resulting in the $x'$ coordinate of other pixels. This procedure is depicted in Fig. \ref{x_interpolation}.
The $y'$-interpolation is performed similarly as shown in Fig. \ref{y_interpolation}. Thereafter, the $(x',y')$ target coordinate of all input pixels are found and each input pixel is known where to be transferred.

\subsubsection{Pixel-to-pixel Warping}

\begin{figure}[!t]
\centering
\includegraphics[width=1.8in]{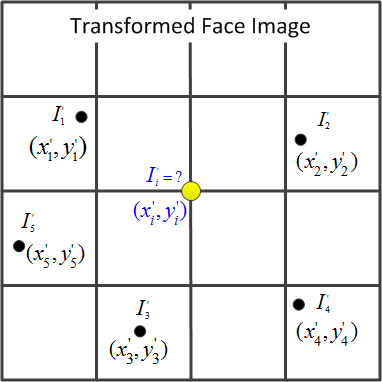}
\caption{intensity-interpolation.}
\label{intensity_interpolation}
\end{figure}

After $x'$- and $y'$-interpolations, each $(x',y')$ coordinate gets the intensity of its corresponding $(x,y)$ from the input face, i.e., 

\begin{equation}
I'(x',y') = I(x,y).
\end{equation}
$I'$ values are then resampled on a uniform grid, e.g., $140 \times 120$ pixels, to create the warped face (see Fig. \ref{intensity_interpolation}). 


\begin{figure}[!t]
\centering
\includegraphics[width=3.4in]{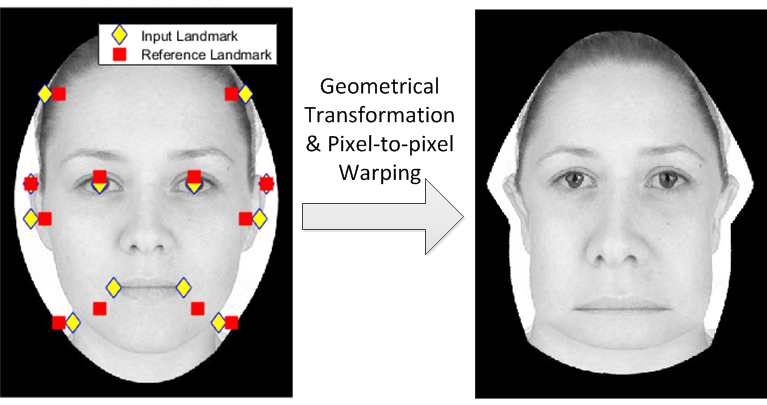}
\caption{An example of geometrical transformation and pixel-to-pixel warping.}
\label{warp_face}
\end{figure} 

\begin{figure}[!t]
\centering
\includegraphics[width=3.4in]{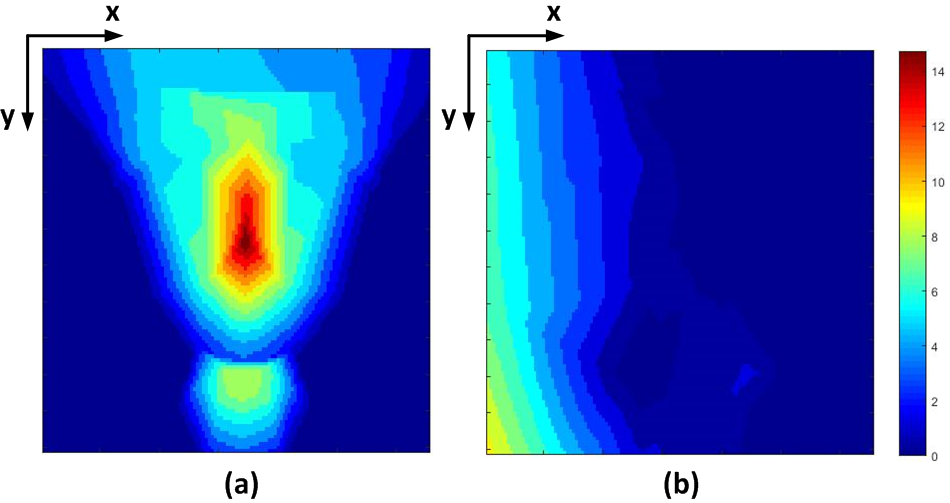}
\caption{Illustration of $\Delta x$ and $\Delta y$ information for a sample warped face. (a) $\Delta x$ information, (b) $\Delta y$ information.}
\label{figure_delta_x_and_y}
\end{figure} 

For the sake of demonstration, an example of geometrical transformation and pixel-to-pixel warping on a sample face with a few number of landmarks is depicted in Fig. \ref{warp_face}.
In this figure, the yellow diamond points and red square points are respectively input and reference landmarks. The face is warped so that the input landmarks are precisely located at the position of reference landmarks, as it was the goal of geometrical transformation. The other pixels are interpolated as explained previously.

\section{Geometrical Information}\label{Geometrical_Information_section}

Geometrical information seems to be useful in addition to intensity information of the warped face.
Obviously, the geometry information of each face exists in its unwarped (input) face. By finding the original coordinate (i.e., coordinate in the unwarped face image) of each pixel of the warped face, geometry information can be gathered.
However, as $x'$ and $y'$ coordinates have been once resampled, their original coordinates cannot be found directly. These coordinates can be obtained by performing two other resamplings on the same grid as before; one for original $x$ values and one for original $y$ values.
To better explain it, two other interpolations are performed in which the $x$ and $y$ source coordinate of each pixel in the warped face is found using interpolation. These two interpolations are exactly the same as previous intensity-interpolation (Fig. \ref{intensity_interpolation}) but by replacing $I'$ with $x$ and $y$.

For the sake of better visualization, the difference of original coordinates $x$ and $y$ of every pixel from its previous pixel is calculated. The differences in original coordinates are denoted as $\Delta x$ and $\Delta y$ here, respectively for differences in $x$ and $y$ information. Figure \ref{figure_delta_x_and_y} illustrates the information of $\Delta x$ and $\Delta y$ for a sample face in Yale dataset \cite{webYaleDataset}. The amount of vertical and horizontal transitions of each pixel after warping can be seen in this figure. This figure shows that for this specific face, warping has changed face more in horizontal direction rather than vertical. 

\section{Classification Using Ensemble of Patches}\label{Classification_section}

\subsection{Ensemble of Patches and Feature Vectors}

Instead of using the whole face, a patch-based approach is used in this work. To do this, an ensemble of patches are created in the limit of face frame. The location of patches are selected randomly once, and for all faces of dataset, the same patches are used in both training and testing phases. 
The optimum number and size of patches were found through trial and error to be 80 and $30 \times 30$ pixels, respectively, over various different datasets.

\begin{figure}[!t]
\centering
\includegraphics[width=3.45in]{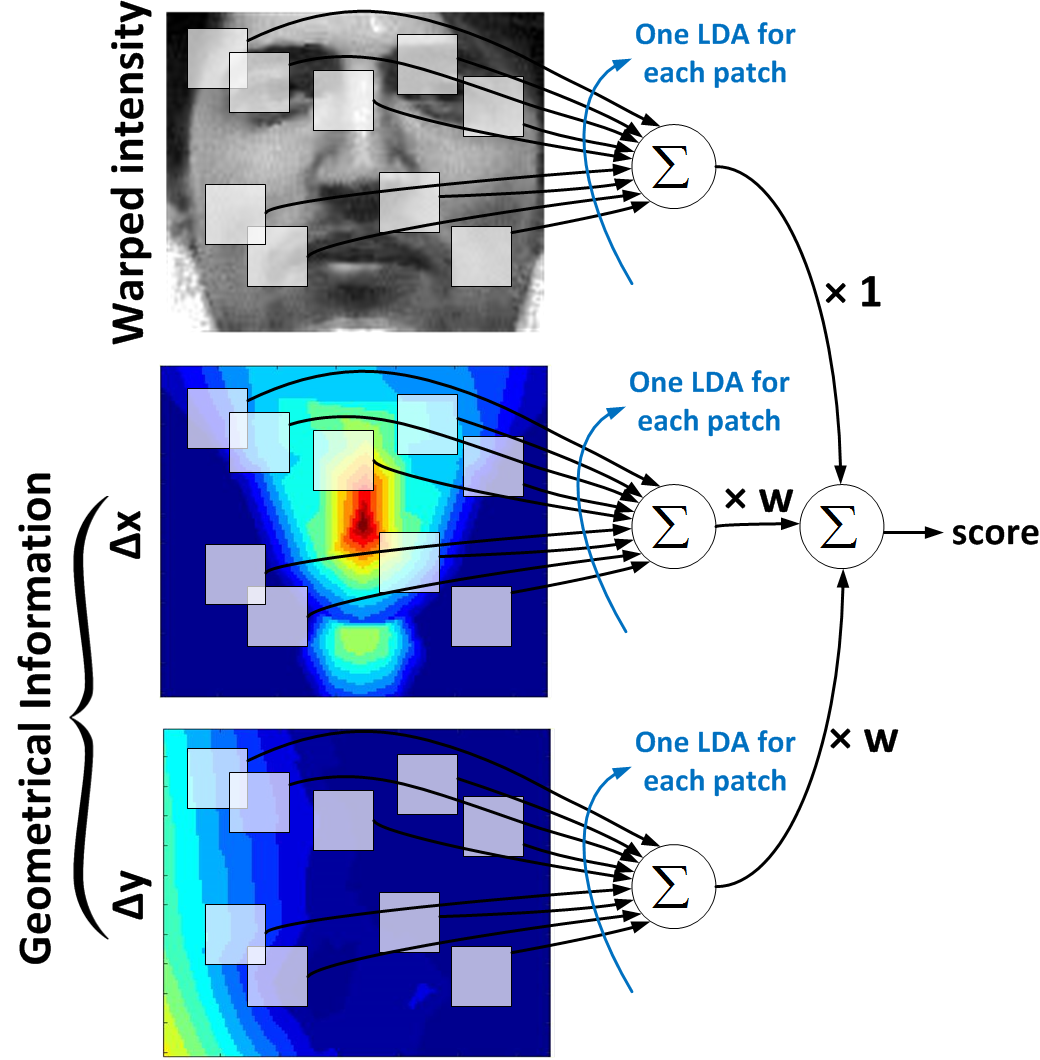}
\caption{Classification using ensemble of patches.}
\label{figure_Patches}
\end{figure}

For every face, the ensemble of patches are applied on intensity matrix of its warped face, its $\Delta x$ information, and its $\Delta y$ matrix. 
An example of applying ensemble of patches on these three matrices is depicted in Fig. \ref{figure_Patches}.
Note that the information of $\Delta x$ and $\Delta y$ is the same as $x$ and $y$.
In order to have the feature vectors of each patch, the matrix coefficients fell in the patch are reshaped as a vector. In other words, for the $p^{th}$ patch, if the size of patch is $m \times m$, the feature vectors are obtained as,
\begin{equation}
f^I_p = [I'(1,1), I'(1,2), \dots, I'(m,m)]^T,
\end{equation}
\begin{equation}
f^{\Delta x}_p = [\Delta x(1,1), \Delta x(1,2), \dots, \Delta x(m,m)]^T,
\end{equation}
\begin{equation}
f^{\Delta y}_p = [\Delta y(1,1), \Delta y(1,2), \dots, \Delta y(m,m)]^T,
\end{equation}
where $f^I_p$, $f^{\Delta x}_p$, and $f^{\Delta y}_p$ are respectively the feature vectors of $p^{th}$ patch with respect to intensity, $\Delta x$, and $\Delta y$ matrices. Moreover, $I'(k,l)$, $\Delta x(k,l)$, and $\Delta y(k,l)$ denote the coefficient of intensity, $\Delta x$, and $\Delta y$ matrices which fall in pixel $(k,l)$ of the patch.

\subsection{Fisher Linear Discriminant Analysis}\label{section_LDA}

After constructing the feature vecotrs of ensemble of patches, three separate Fisher Linear Discriminant Analysis (LDA) subspaces are trained for every patch. To better explain, for $p^{th}$ patch in all training set of faces, one Fisher LDA subspace is trained using the feature vectors $f^I_p$, one for feature vectors $f^{\Delta x}_p$, and one for feature vectors $f^{\Delta y}_p$.
In this work, Fisherface method \cite{belhumeur1997eigenfaces} is used for classification of each patch; however, other more complicated learning methods can be used in future works.

The goal of Fisher LDA is maximizing the ratio of,

\begin{equation}
\label{Fisher_optimize_equation}
W_{opt} = \text{arg }\underset{W}{\text{max}} \frac{|W^T S_b W|}{|W^T S_w W|},
\end{equation}
where $S_b$ and $S_w$ are the between- and within-class scattering matrices, respectively \cite{hastie2002elements,bishop2007pattern}, formulated as,

\begin{equation}
S_w = \sum_{i=1}^C \sum_{x_k \in X_i} (x_k - \mu_i) (x_k - \mu_i)^T,
\end{equation}

\begin{equation}
S_b = \sum_{i=1}^C N_i (\mu_i - \mu) (\mu_i - \mu)^T,
\end{equation}
where $\mu_i$ is the mean of $i^{th}$ class and $\mu$ is the mean of means of classes. $N_i$ is the number of samples of $i^{th}$ class. And $x_k$ is the $k^{th}$ sample of $i^{th}$ class ($X_i$).   

After finding scattering matrices, a discriminative subspace is created using the eigenvectors of $S_w^{-1}S_b$ matrix. 
To extract the discriminative features from each feature vector, it should be projected onto this subspace.
If $C$ denotes the number of classes, this projection also reduces the dimension of data to $C-1$ \cite{hastie2002elements,bishop2007pattern}.

\begin{figure*}[!t]
\centering
\includegraphics[width=4.5in]{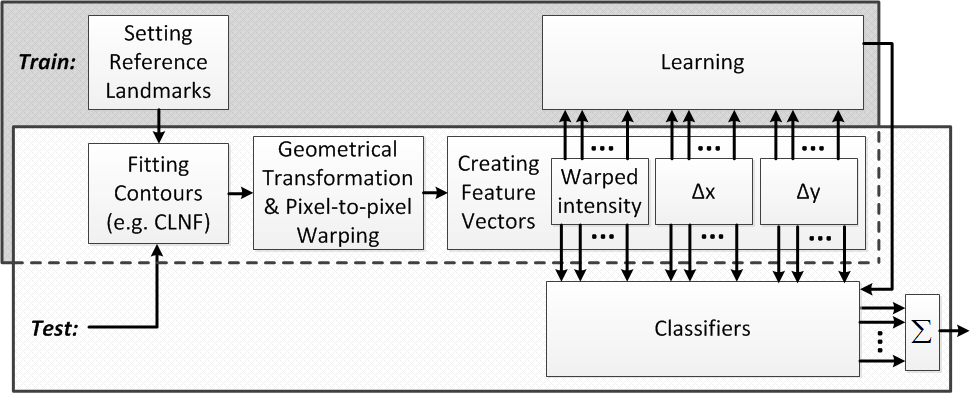}
\caption{The overall structure of the proposed method.}
\label{final_structure}
\end{figure*}

\subsection{Decision Making}

Clearly, there are a lot of different features available rather than one, i.e., intensity, $\Delta x$, and $\Delta y$ features for all patches. Hence, in order to obtain the final similarity/distance score between two face images, a fusion is required to be performed. The fusion can be performed either before, during, or after classification, which are respectively known as data-, feature-, and decision-level fusion.
In the fusion of data and feature, respectively, the two feature vectors are concatenated before and after projecting to the discriminative subspace; and in the fusion of decision, the resulting scores are fused. The fusion of decision is found to perform better in this work. 

For $p^{th}$ patch in every face image, each of the feature vectors, $f^I_p$, $f^{\Delta x}_p$, and $f^{\Delta y}_p$, is projected onto their corresponding discriminative LDA subspace, obtained as described in Section \ref{section_LDA}.
The projections result in projected feature vectors $\widehat{f}^I_p$, $\widehat{f}^{\Delta x}_p$, and $\widehat{f}^{\Delta y}_p$.
In the context of face recognition, it has been shown that the cosine of the angle between two discriminative feature vectors, which is obviously a similarity score, results a better recognition rather than distance measures such as Euclidean distance \cite{perlibakas2004distance,mohammadzade2013projection}. Hence, cosine is used in this work for matching purposes. 
Then, the similarity score between two face images $i$ and $j$ is calculated as follows. First, the similarity scores in the discriminative subspaces related to $p^{th}$ patch are obtained as,

\begin{equation}
\text{sim}^I_p (i,j) = \text{cos}(\widehat{f}_{p,i}^{I},\widehat{f}_{p,j}^{I}) = \frac{\widehat{f}_{p,i}^{I} \widehat{f}_{p,j}^{I}}{|\widehat{f}_{p,i}^{I}| |\widehat{f}_{p,j}^{I}|},
\end{equation}
\begin{equation}
\text{sim}^{\Delta x}_p (i,j) = \text{cos}(\widehat{f}_{p,i}^{\Delta x},\widehat{f}_{p,j}^{\Delta x}) = \frac{\widehat{f}_{p,i}^{\Delta x} \widehat{f}_{p,j}^{\Delta x}}{|\widehat{f}_{p,i}^{\Delta x}| |\widehat{f}_{p,j}^{\Delta x}|},
\end{equation}
\begin{equation}
\text{sim}^{\Delta y}_p (i,j) = \text{cos}(\widehat{f}_{p,i}^{\Delta y},\widehat{f}_{p,j}^{\Delta y}) = \frac{\widehat{f}_{p,i}^{\Delta y} \widehat{f}_{p,j}^{\Delta y}}{|\widehat{f}_{p,i}^{\Delta y}| |\widehat{f}_{p,j}^{\Delta y}|},
\end{equation}
where $\widehat{f}_{p,i}^{I}$, $\widehat{f}_{p,i}^{\Delta x}$, and $\widehat{f}_{p,i}^{\Delta y}$ are respectively projected feature vectors $\widehat{f}^I_p$, $\widehat{f}^{\Delta x}_p$, and $\widehat{f}^{\Delta y}_p$ in $i^{th}$ face image.

Then, the final similarity score is simply obtained by a weighted summation of all the scores of patches (decision fusion),
\begin{equation}
\text{sim} (i,j) = \sum_{p=1}^{80} (\text{sim}^I_p (i,j) + w \text{ sim}^{\Delta x}_p (i,j) + w \text{ sim}^{\Delta y}_p (i,j) ),
\end{equation}
where $w$ is the weight associated to the geometrical information, and the weight of intensity information is considered to be one for simplicity. The classification using ensemble of patches is summarized in Fig. \ref{figure_Patches}.

\subsection{Overall structure of the proposed face recognition framework}

The proposed method can be summarized as is depicted in Fig. \ref{final_structure}. In this method, a set of reference contours is constructed, landmarks of each train/test face are detected using CLNF method, the faces are transformed geometrically to the reference, warping is performed, and feature vectors are created for classification. In preparing feature vectors, the ensemble of patches are considered for matrices of warped intensity, $\Delta x$, and $\Delta y$. A separate Fisher LDA is trained for every patch in each of these matrices. Finally, in the test phase, the feature vectors are projected onto the corresponding LDA subspaces and the similarity scores are summed up together in order to have the total score.

\section{Experimental Results}\label{Experimental_results_section}

\subsection{Datasets}

Four different datasets are used for evaluating the recognition performance using the proposed alignment method, which are Yale \cite{webYaleDataset}, AT\&T \cite{webAttDataset}, Cohn-Kanade \cite{webCohnDataset,kanade2000comprehensive}, and LFW datasets \cite{webLfwDataset,huang2007labeled} detailed in the following. In this work, as a pre-processing, the datasets are eye-aligned and then cropped using CLNF method \cite{baltrusaitis2013constrained}. To more explain, the location of eyes are found using CLNF and by a translation and rotation, the faces become eye-aligned.

\subsubsection{Yale face dataset}

The Yale face dataset \cite{webYaleDataset} was created by the Center of Computational Vision and Control at Yale University, New Haven. It consists of 165 grayscale face images of 15 different persons. There exist 11 images per person depicting different facial expressions.

\subsubsection{AT\&T face dataset}

The AT\&T face dataset \cite{webAttDataset} was created by the AT\&T Laboratories Cambridge in 2002. There are pictures of 40 different persons with 10 different facial expressions.

\begin{figure*}[!t]
\centering
\includegraphics[width=6in]{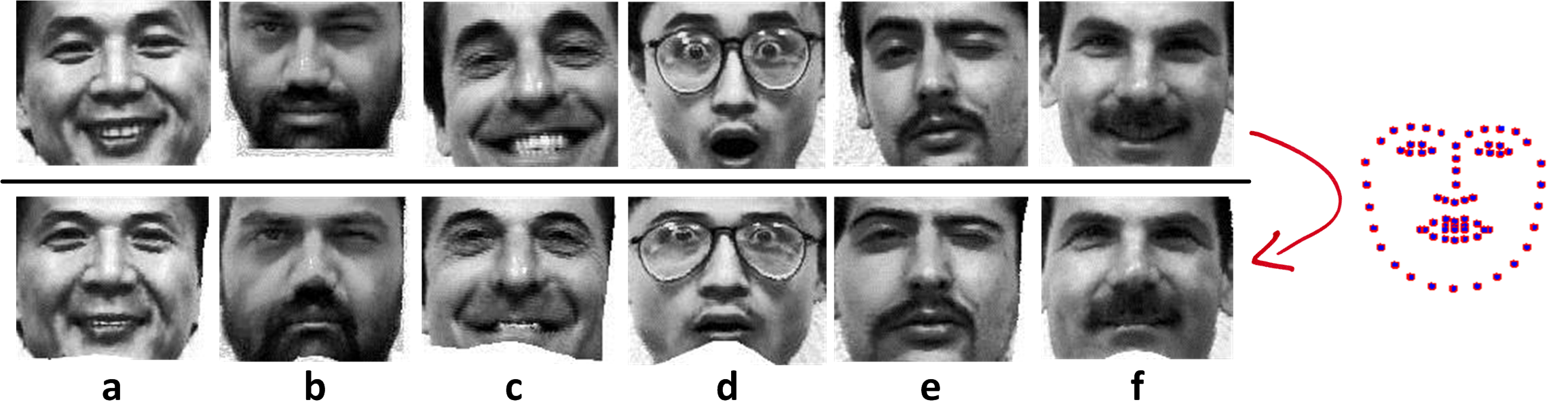}
\caption{Several samples of pixel alignment in Yale dataset}
\label{Warp_Yale}
\end{figure*}

\begin{figure*}[!t]
\centering
\includegraphics[width=6in]{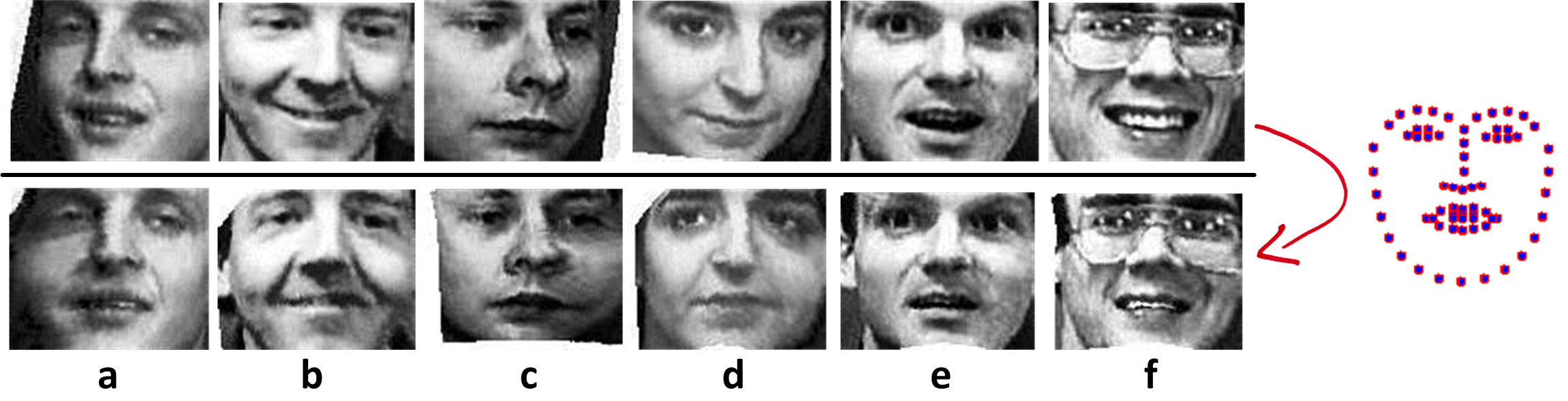}
\caption{Several samples of pixel alignment in AT\&T dataset}
\label{Warp_ATT}
\end{figure*}

\subsubsection{Cohn-Kanade face dataset}

Cohn-Kanade dataset \cite{webCohnDataset,kanade2000comprehensive} includes 486 face sequences from 97 persons. Every sequence starts with neutral face and ends with extreme versions of expressions. Different expressions exist in this dataset, such as laughing, surprising, and etc. The first version of this dataset is used here. For every person in this dataset, merely one neutral face, all middle expressions, and all extreme expressions are utilized in this work to perform experiments. 

\subsubsection{LFW face dataset}

Labeled Faces in the Wild (LFW) dataset \cite{webLfwDataset,huang2007labeled} is a very big and challenging dataset including 13,233 images of faces collected from the web. The faces have various poses, expressions, and locations in images. The distances of camera from persons are not necessarily the same in images. There are different number of images for every subject, from one to sometimes 10. The not-cropped and not-processed version of this dataset is used in this work for experiments.

\subsection{Warped Faces}

In this section, for the sake of visualization, several warped faces which are pixel aligned are shown and analyzed. Several samples of warped faces from Yale and AT\&T datasets are illustrated in figures \ref{Warp_Yale} and \ref{Warp_ATT}, respectively. In these figures, the first and second row are faces before and after warping, respectively. At the right-hand side of these figures, the reference contours are shown.

As seen in Fig. \ref{Warp_Yale}, faces (a), (c) and (f) are smiling originally but the mouths in the corresponding warped faces are closed and the teeth are roughly removed. Face (d), however, is wondering originally while the mouth is totally closed after warping. 
Similarly, in Fig. \ref{Warp_ATT}, faces (b), (d), (e) and (f) have different expressions while their corresponding warped faces have neutral expression with closed mouths.
As shown in these figures, removing the facial expression is obviously one of the results of the proposed pixel-by-pixel alignment method, which of course can greatly improve the recognition task.
Moreover, faces (b), (c) and (e) in Fig. \ref{Warp_Yale} show that this method can also change the pose of faces to the pose of the reference contours. 
Similarly, in Fig. \ref{Warp_ATT}, faces (a), (b), (c) and (d) have frontal pose after warping.
Clearly, in all the warped faces in figures \ref{Warp_Yale} and \ref{Warp_ATT}, not only are different organs of the face aligned, but also other features of the face are almost aligned. However, due to the drawback of the landmark detection method in converging to exact landmark points, some features may not become well-aligned. For instance, in Fig. \ref{Warp_Yale}, the eyes are not completely open in the warped faces (a), (b), (c), (e) and (f).

\subsection{Experiments}

In all the experiments mentioned in this section, the dataset is firstly shuffled randomly and then 5-fold cross validation is performed. In the following, experimenting the impact of patch size are reported and analyzed. Thereafter, classification using ensemble of patches is compared to classification using the whole face. Finally, the proposed method is examined and compared to eye-aligned classification.

\subsubsection{Experiment on Size of Patches}

In this experiment, the effect of patch sizes are mentioned and reported. Different experiments on AT\&T dataset were performed with different sizes of patches, which are $10 \times 10$, $20 \times 20$, $30 \times 30$, $40 \times 40$, $50 \times 50$, and random-sized patches each with one of the mentioned sizes. In these experiments, 80 random patches were utilized, and the optimum weight of geometrical information was found to be 0.2 through trial and error.

\begin{figure}[!t]
\centering
\includegraphics[width=3.45in]{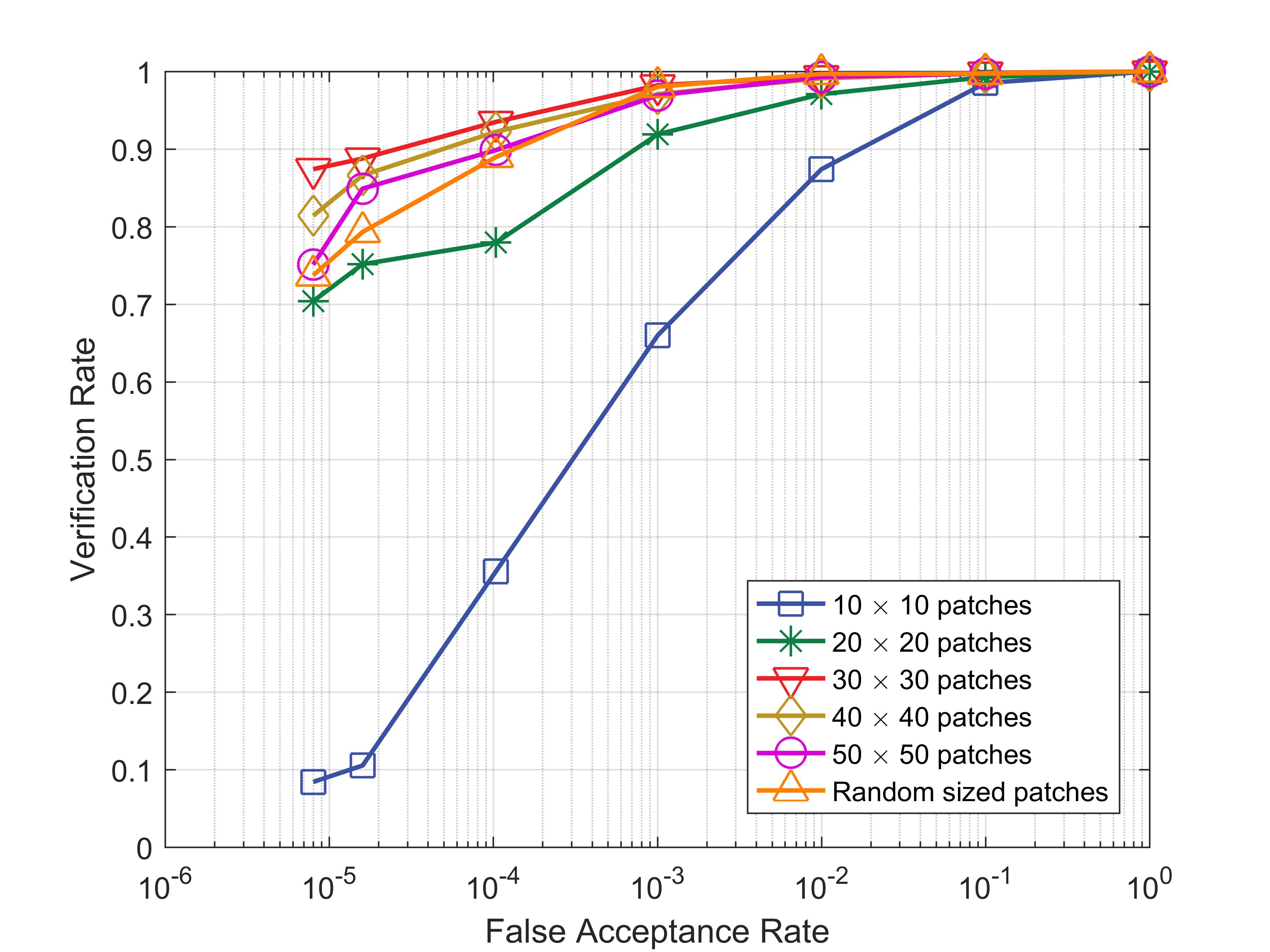}
\caption{Effect of size of patches in classification using ensemble of patches}
\label{EXP_size_patch_ATT}
\end{figure}

In each iteration of the experiments, the similarity score between every pair of gallery and probe images is calculated and the Receiver Operating Characteristic (ROC) curve using all the scores are plotted. The ROC curves of experiments are depicted in Fig. \ref{EXP_size_patch_ATT}. As is obvious in this figure, the size of patches has important impact on the recognition rate. According to the curves, $30 \times 30$ patches perform better; therefore,  merely $30 \times 30$ patches are used in the next experiments.

\subsubsection{Patch-based Recognition Using Eye-aligned and Pixel-aligned Faces}\label{section_wholeVsPatch}

\begin{figure}[!t]
\centering
\includegraphics[width=3.45in]{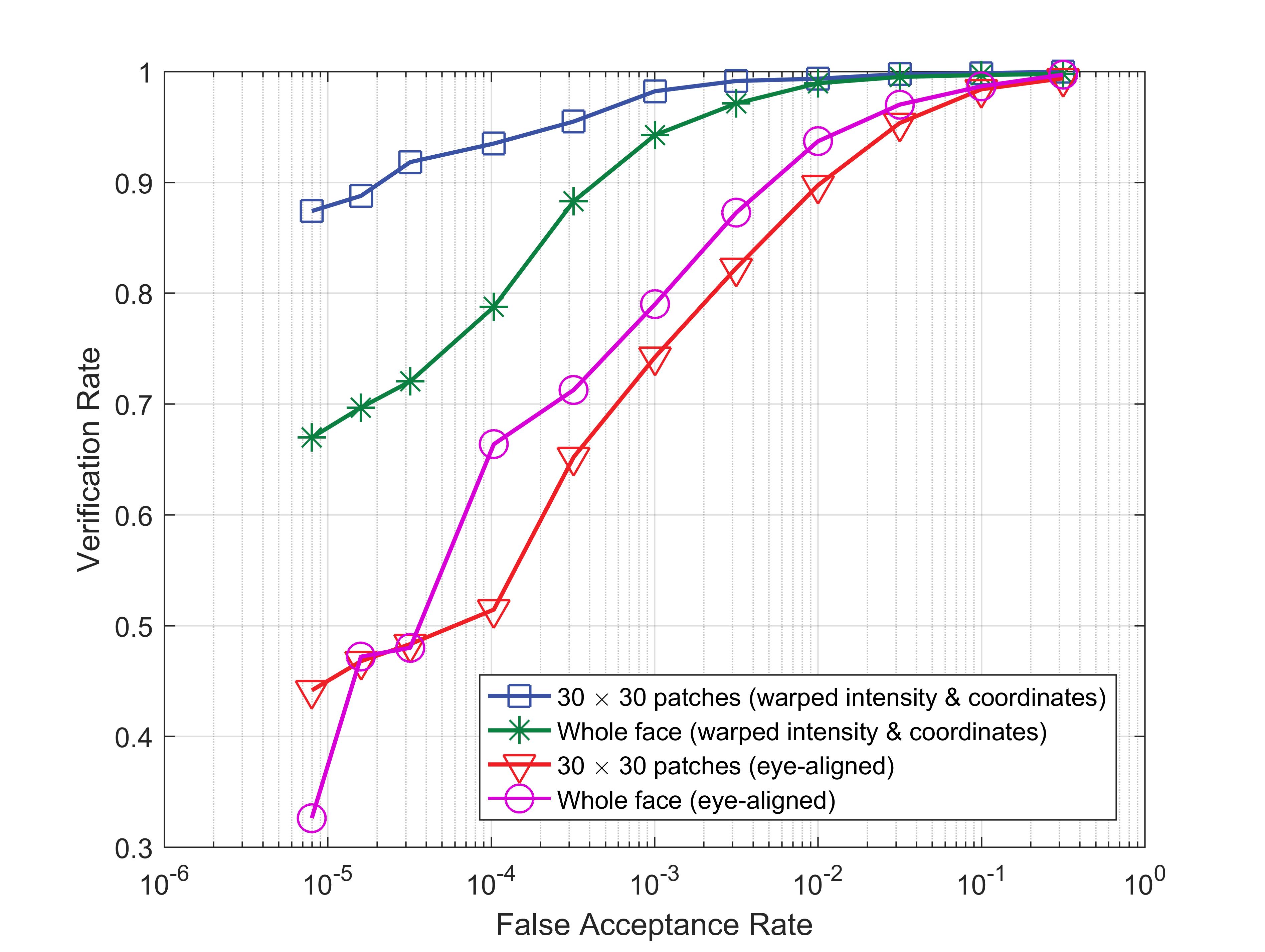}
\caption{Comparison of classification using the whole face or ensemble of patches}
\label{EXP_whole_versus_patch_ATT}
\end{figure}

Several other experiments were performed evaluating the effect of using ensemble of patches for both eye-aligned and pixel-aligned face images. In these experiments, 80 random patches were utilized with size $30 \times 30$. 
First, classification using ensemble of patches and not using patches were tested on eye-aligned images. 
Note that in classification using ensemble of patches for eye-aligned faces, the ensemble was solely applied on intensity matrix of eye-aligned images because warping does not exist anymore and thus there is no geometrical information.
Figure \ref{EXP_whole_versus_patch_ATT} shows ROC curves of the two experiments performed on AT\&T dataset.
As can be seen in this figure, using ensemble of patches results in overall worse performance than not using patches when eye-aligned method is utilized.

On the other hand, the same two experiments were performed using pixel-aligned faces rather than eye-aligned ones. The ROC curves of these experiments on AT\&T dataset are also depicted in Fig. \ref{EXP_whole_versus_patch_ATT}. As obvious in this figure, when pixel-aligned faces are used, patch-based recognition produces superior results compared to not using patches. 
For instance, in FAR of 0.001, verification rates are roughly 99\% and 94\% in recognition using ensemble of patches and not using it, respectively. This result verifies the effectiveness of using ensemble of patches alongside having faces pixel-to-pixel warped.

\subsubsection{Eye-aligned Versus Proposed Method}

\begin{figure*}[!t]
\centering
\includegraphics[width=6.5in]{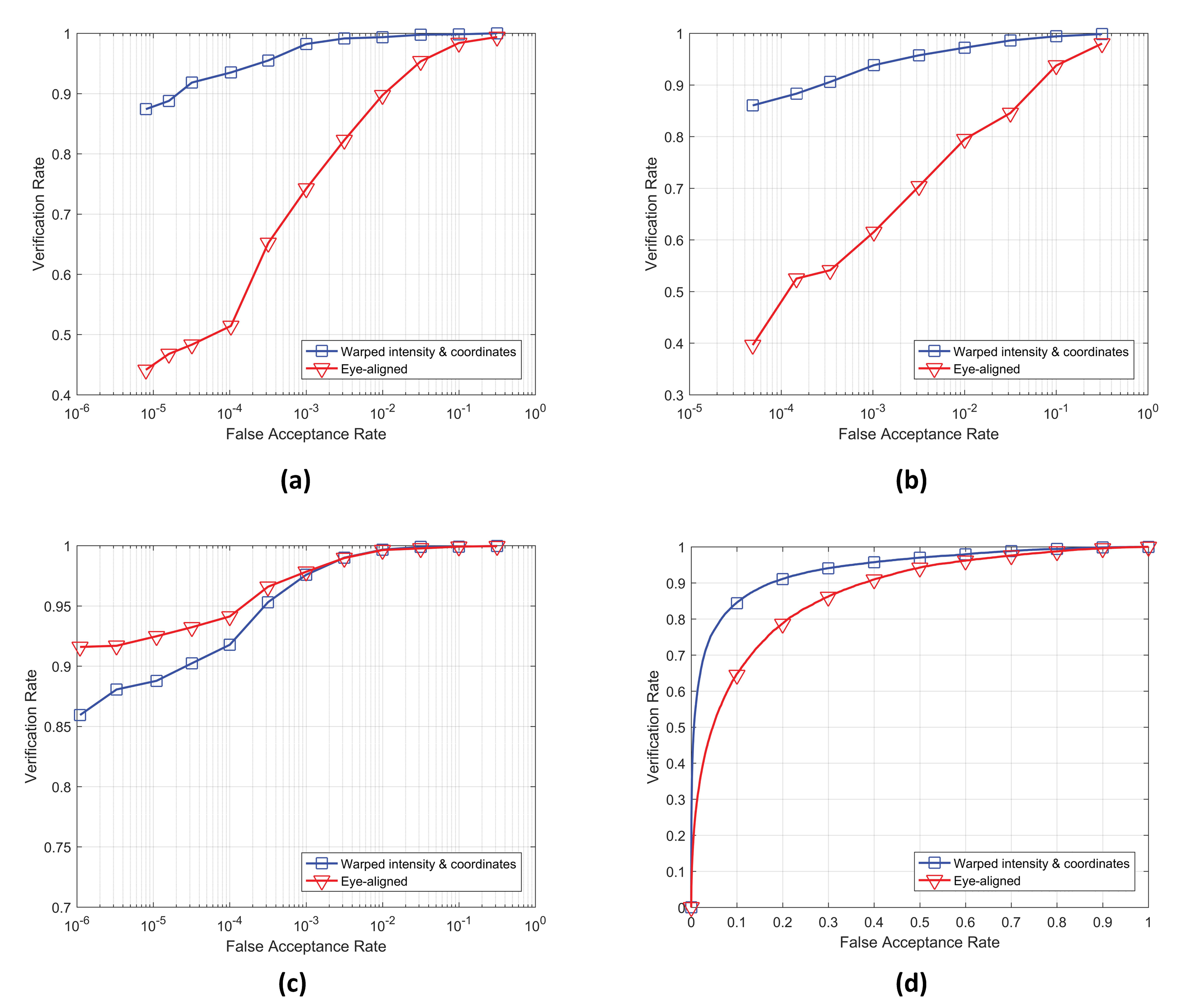}
\caption{Comparison of proposed method with eye-aligned face recognition on (a) Yale dataset, (b) AT\&T dataset, (c) Cohn-Kanade dataset, (d) LFW dataset.}
\label{EXP_eyeAlignVersusWarp}
\end{figure*}

Eye-aligned face recognition is compared with the proposed pixel-aligned classification method in Fig. \ref{EXP_eyeAlignVersusWarp} and Table \ref{table_results}. This comparison is performed for four datasets, which are Yale \cite{webYaleDataset}, AT\&T \cite{webAttDataset}, Cohn-Kanade \cite{webCohnDataset,kanade2000comprehensive}, and LFW datasets \cite{webLfwDataset,huang2007labeled}. LFW dataset is a very challenging and big dataset and includes images which might have more than one face, but still there is only one subject label associated with each image. For this dataset, using CLNF method \cite{baltrusaitis2013constrained}, the faces in image were detected. If there were several detected faces in the image, the face with the biggest area (multiplication of height and width of face) and minimum distance from the center of image was extracted as the main face. Thereafter, the main face was cropped out of the image.

\begin{table}[!t]
\renewcommand{\arraystretch}{1.3}  
\caption{Results of the proposed method and eye-aligned face recognition in specific false alarm rates.}
\label{table_results}
\centering
\begin{threeparttable}[b]
\begin{tabular}{l || c | c | c}
\hline
\hline
\textbf{Dataset} & \textbf{Image} & \textbf{FAR\tnote{1}} & \textbf{Verification Rate} \\
\hline
\hline
\multirow{2}{*}{Yale}  
& \multirow{1}{*}{Eye-aligned} & 0.001 & 74\% \\
\cline{2-4}
& \multirow{1}{*}{Proposed alignment} & 0.001 & 99\% \\
\hline
\multirow{2}{*}{AT\&T}  
& \multirow{1}{*}{Eye-aligned} & 0.001 & 62\% \\
\cline{2-4}
& \multirow{1}{*}{Proposed alignment} & 0.001 & 94\% \\
\hline
\multirow{2}{*}{Cohn-Kanade}  
& \multirow{1}{*}{Eye-aligned} & 0.001 & 95.6\% \\
\cline{2-4}
& \multirow{1}{*}{Proposed alignment} & 0.001 & 95.5\% \\
\hline
\multirow{2}{*}{LFW}  
& \multirow{1}{*}{Eye-aligned} & 0.1 & 64\% \\
\cline{2-4}
& \multirow{1}{*}{Proposed alignment} & 0.1 & 84\% \\
\hline
\end{tabular}
\begin{tablenotes}
	\item [1] The reason of choosing False Acceptance Rate (FAR) of 0.1, rather than 0.001, for LFW dataset is that this dataset is very challenging giving sense to this rather easier false acceptance rate.
\end{tablenotes}
\end{threeparttable}
\end{table}

As can be seen in the ROC curves of Fig. \ref{EXP_eyeAlignVersusWarp} and Table \ref{table_results}, the proposed method significantly outperforms eye-aligned face recognition with a wonderful enhancement in Yale, AT\&T, and LFW dataset. Notice that LFW is a very challenging and big dataset, and Yale and AT\&T datasets are two medium well-known datasets. The proposed method results very good in both big and small datasets, showing its power and effectiveness in different types of datasets.

In Cohn-Kanade dataset, however, eye-aligned method performs slightly better than the proposed method; although the ROC curves show that the rates of proposed method is almost near the rate of eye-aligned face recognition. The reason of this failure is that the CLNF method \cite{baltrusaitis2013constrained}, which was used for warping, did not work precisely in detecting very open mouths in extremely surprising faces. Thus, warping could not be performed successfully because of imprecise detected landmarks. Therefore, this failure is not because of the weakness of the proposed method, but because of not having correct and accurate landmarks as input. 

\section{Discussions}\label{Discussions_section}

\subsection{Discussion on Alignment of Features}

The most important contribution of this work was to introduce a method for fine alignment of intensity features of the face which as a by-product also results in accurate extraction of geometry information of the face. In other words, the proposed alignment method places the intensity of similar organs in the same positions in the warped faces. On the other hand, when intensities are properly aligned, using the proposed geometry extraction method, the coordinates of the aligned pixels can be extracted as the corresponding geometry information. As a result, pixel alignment provides both intensity and geometry information useful for recognition.
Moreover, it is important to note that the key finding of this work, i.e., finer alignment results in better classification, is not limited to face recognition problem and is applicable to any other pattern recognition/classification challenge where features are badly corresponded. However, in such pattern recognition problems, what is required to be considered first is to construct an alignment framework, i.e., defining proper correspondences and creating a method for aligning to a reference assortment of features.

\subsection{Discussion on Classification Using Ensemble of Patches}

As it was experimented in Section \ref{section_wholeVsPatch} and shown in Fig. \ref{EXP_whole_versus_patch_ATT}, it was observed that for warped faces, classification using ensemble of patches enhances performance in comparison to classification using the whole face. However, this enhancement is not seen for eye-aligned (not warped) faces. This might be because of the fact that in warped faces, every pixel corresponds to a specific region (such as lip corners) in all faces; however, it is not true in eye-aligned images. Therefore, in warped faces, every patch covers similar and corresponding pixels in all faces but may cover not related pixels in eye-aligned ones. That is why using patches has made the result worse in eye-aligned faces, as well as improving result in warped faces.

\section{Conclusion}\label{Conclusion_section}

In this article, a pixel-level facial alignment method, i.e., a method to align the whole pixels of faces is proposed. This alignment is achieved by mapping the face geometry onto a reference geometry, where the mapping is guided by contours of facial landmarks which are fitted to each face using landmark detection methods such as CLNF \cite{baltrusaitis2013constrained}. 
The proposed alignment method provides both the aligned intensity information and their corresponding geometry information. 
The resulting aligned intensity and geometry features create superior recognition results when they are used in a patch-based recognition framework.   

The experiments were performed on four well-known datasets and Fisherfaces \cite{belhumeur1997eigenfaces} was used as an instance of a holistic-based face recognition method. Results showed the significantly better performance of patch-based pixel-aligned face recognition in comparison to eye-aligned face recognition in all utilized datasets (except on Cohn-Kanade dataset with slightly rate difference). The reason of not having outperformance in this dataset is that the landmark detection, which is not a contribution of this work, did not work properly in extreme expressions resulting in not qualified warping. The proposed method guarantees better performance in comparison to eye-aligned face recognition when the landmarks are detected properly. 

\section{Future Work}\label{Future_work_section}

In the proposed warping, all faces are warped to a unique neutral face from different expressions and poses. When the geometrical information is obtained using the warped and original faces, three different type of information are included in it, i.e., the face itself, expression and pose. Among these three pieces of information, merely the face itself is important for us because it reflects which pixel has gone where. The other two ones, which are expression and pose, make geometrical information impure because two different expressions or poses of one person result in different geometrical information which is not good. One solution to this problem is not to have only one reference face, but to have one reference face per every expression or pose. This can be performed using regression for every expression or pose with landmarks of non-neutral face as input and landmarks of neutral image as output. We are looking to it as a future work.


%

\appendices




\ifCLASSOPTIONcaptionsoff
  \newpage
\fi



%

\bibliographystyle{IEEEtran}
\bibliography{References}

\begin{thebibliography}{10}
\providecommand{\url}[1]{#1}
\csname url@samestyle\endcsname
\providecommand{\newblock}{\relax}
\providecommand{\bibinfo}[2]{#2}
\providecommand{\BIBentrySTDinterwordspacing}{\spaceskip=0pt\relax}
\providecommand{\BIBentryALTinterwordstretchfactor}{4}
\providecommand{\BIBentryALTinterwordspacing}{\spaceskip=\fontdimen2\font plus
\BIBentryALTinterwordstretchfactor\fontdimen3\font minus
  \fontdimen4\font\relax}
\providecommand{\BIBforeignlanguage}[2]{{%
\expandafter\ifx\csname l@#1\endcsname\relax
\typeout{** WARNING: IEEEtran.bst: No hyphenation pattern has been}%
\typeout{** loaded for the language `#1'. Using the pattern for}%
\typeout{** the default language instead.}%
\else
\language=\csname l@#1\endcsname
\fi
#2}}
\providecommand{\BIBdecl}{\relax}
\BIBdecl

\bibitem{zhao2003face}
W.~Zhao, R.~Chellappa, P.~J. Phillips, and A.~Rosenfeld, ``Face recognition: A
  literature survey,'' \emph{ACM computing surveys (CSUR)}, vol.~35, no.~4, pp.
  399--458, 2003.

\bibitem{kelly1970visual}
M.~D. Kelly, ``Visual identification of people by computer.'' STANFORD UNIV
  CALIF DEPT OF COMPUTER SCIENCE, Tech. Rep., 1970.

\bibitem{kanade1977computer}
T.~Kanade, \emph{Computer recognition of human faces}.\hskip 1em plus 0.5em
  minus 0.4em\relax Birkh{\"a}user Basel, 1977, vol.~47.

\bibitem{brunelli1993face}
R.~Brunelli and T.~Poggio, ``Face recognition: Features versus templates,''
  \emph{IEEE transactions on pattern analysis and machine intelligence},
  vol.~15, no.~10, pp. 1042--1052, 1993.

\bibitem{nefian1998hidden}
A.~V. Nefian and M.~H. Hayes, ``Hidden markov models for face recognition,'' in
  \emph{Acoustics, Speech and Signal Processing, 1998. Proceedings of the 1998
  IEEE International Conference on}, vol.~5.\hskip 1em plus 0.5em minus
  0.4em\relax IEEE, 1998, pp. 2721--2724.

\bibitem{samaria1994hmm}
F.~Samaria and S.~Young, ``Hmm-based architecture for face identification,''
  \emph{Image and vision computing}, vol.~12, no.~8, pp. 537--543, 1994.

\bibitem{ding2015multi}
C.~Ding, C.~Xu, and D.~Tao, ``Multi-task pose-invariant face recognition,''
  \emph{IEEE Transactions on Image Processing}, vol.~24, no.~3, pp. 980--993,
  2015.

\bibitem{chen2013blessing}
D.~Chen, X.~Cao, F.~Wen, and J.~Sun, ``Blessing of dimensionality:
  High-dimensional feature and its efficient compression for face
  verification,'' in \emph{Proceedings of the IEEE Conference on Computer
  Vision and Pattern Recognition}, 2013, pp. 3025--3032.

\bibitem{pentland1994view}
A.~Pentland, B.~Moghaddam, T.~Starner \emph{et~al.}, ``View-based and modular
  eigenspaces for face recognition,'' in \emph{CVPR}, vol.~94, 1994, pp.
  84--91.

\bibitem{penev1996local}
P.~S. Penev and J.~J. Atick, ``Local feature analysis: A general statistical
  theory for object representation,'' \emph{Network: computation in neural
  systems}, vol.~7, no.~3, pp. 477--500, 1996.

\bibitem{cootes1998active}
T.~F. Cootes, G.~J. Edwards, and C.~J. Taylor, ``Active appearance models,'' in
  \emph{Computer Vision, 1998. Proceedings of the 1998 European Conference on},
  vol.~2.\hskip 1em plus 0.5em minus 0.4em\relax Springer, 1998, pp. 484--498.

\bibitem{cootes2001active}
------, ``Active appearance models,'' \emph{IEEE Transactions on pattern
  analysis and machine intelligence}, vol.~23, no.~6, pp. 681--685, 2001.

\bibitem{lanitis1995automatic}
A.~Lanitis, C.~J. Taylor, and T.~F. Cootes, ``Automatic face identification
  system using flexible appearance models,'' \emph{Image and vision computing},
  vol.~13, no.~5, pp. 393--401, 1995.

\bibitem{stegmann2002analysis}
M.~B. Stegmann, ``Analysis and segmentation of face images using point
  annotations and linear subspace techniques,'' Tech. Rep., 2002.

\bibitem{edwards1998face}
G.~J. Edwards, T.~F. Cootes, and C.~J. Taylor, ``Face recognition using active
  appearance models,'' in \emph{European conference on computer vision}.\hskip
  1em plus 0.5em minus 0.4em\relax Springer, 1998, pp. 581--595.

\bibitem{lanitis1995unified}
A.~Lanitis, C.~J. Taylor, and T.~F. Cootes, ``A unified approach to coding and
  interpreting face images,'' in \emph{Computer Vision, 1995. Proceedings.,
  Fifth International Conference on}.\hskip 1em plus 0.5em minus 0.4em\relax
  IEEE, 1995, pp. 368--373.

\bibitem{cootes1995active}
T.~F. Cootes, C.~J. Taylor, D.~H. Cooper, and J.~Graham, ``Active shape
  models-their training and application,'' \emph{Computer vision and image
  understanding}, vol.~61, no.~1, pp. 38--59, 1995.

\bibitem{craw1992face}
I.~Craw and P.~Cameron, ``Face recognition by computer.'' in \emph{BMVC}, 1992,
  pp. 1--10.

\bibitem{turk1991eigenfaces}
M.~Turk and A.~Pentland, ``Eigenfaces for recognition,'' \emph{Journal of
  cognitive neuroscience}, vol.~3, no.~1, pp. 71--86, 1991.

\bibitem{turk1991face}
M.~A. Turk and A.~P. Pentland, ``Face recognition using eigenfaces,'' in
  \emph{Computer Vision and Pattern Recognition, 1991. Proceedings CVPR'91.,
  IEEE Computer Society Conference on}.\hskip 1em plus 0.5em minus 0.4em\relax
  IEEE, 1991, pp. 586--591.

\bibitem{belhumeur1997eigenfaces}
P.~N. Belhumeur, J.~P. Hespanha, and D.~J. Kriegman, ``Eigenfaces vs.
  fisherfaces: Recognition using class specific linear projection,'' \emph{IEEE
  Transactions on pattern analysis and machine intelligence}, vol.~19, no.~7,
  pp. 711--720, 1997.

\bibitem{lu2003face}
J.~Lu, K.~N. Plataniotis, and A.~N. Venetsanopoulos, ``Face recognition using
  kernel direct discriminant analysis algorithms,'' \emph{IEEE Transactions on
  Neural Networks}, vol.~14, no.~1, pp. 117--126, 2003.

\bibitem{lu2005kernel}
J.~Lu, K.~Plataniotis, and A.~Venetsanopoulos, ``Kernel discriminant learning
  with application to face recognition,'' in \emph{Support Vector Machines:
  Theory and Applications}.\hskip 1em plus 0.5em minus 0.4em\relax Springer,
  2005, pp. 275--296.

\bibitem{phillips1999support}
P.~J. Phillips, ``Support vector machines applied to face recognition,'' in
  \emph{Advances in Neural Information Processing Systems}, 1999, pp. 803--809.

\bibitem{moghaddam2000bayesian}
B.~Moghaddam, T.~Jebara, and A.~Pentland, ``Bayesian face recognition,''
  \emph{Pattern Recognition}, vol.~33, no.~11, pp. 1771--1782, 2000.

\bibitem{kasar2016face}
M.~M. Kasar, D.~Bhattacharyya, and T.-h. Kim, ``Face recognition using neural
  network: a review,'' \emph{International Journal of Security and Its
  Applications}, vol.~10, no.~3, pp. 81--100, 2016.

\bibitem{lin1997face}
S.-H. Lin, S.-Y. Kung, and L.-J. Lin, ``Face recognition/detection by
  probabilistic decision-based neural network,'' \emph{IEEE transactions on
  neural networks}, vol.~8, no.~1, pp. 114--132, 1997.

\bibitem{simon2016improved}
M.~O. Sim{\'o}n, C.~Corneanu, K.~Nasrollahi, O.~Nikisins, S.~Escalera, Y.~Sun,
  H.~Li, Z.~Sun, T.~B. Moeslund, and M.~Greitans, ``Improved rgb-dt based face
  recognition,'' \emph{Iet Biometrics}, vol.~5, no.~4, pp. 297--303, 2016.

\bibitem{abdalmageed2016face}
W.~AbdAlmageed, Y.~Wu, S.~Rawls, S.~Harel, T.~Hassner, I.~Masi, J.~Choi,
  J.~Lekust, J.~Kim, P.~Natarajan \emph{et~al.}, ``Face recognition using deep
  multi-pose representations,'' in \emph{Applications of Computer Vision
  (WACV), 2016 IEEE Winter Conference on}.\hskip 1em plus 0.5em minus
  0.4em\relax IEEE, 2016, pp. 1--9.

\bibitem{parkhi2015deep}
O.~M. Parkhi, A.~Vedaldi, A.~Zisserman \emph{et~al.}, ``Deep face
  recognition.'' in \emph{BMVC}, vol.~1, no.~3, 2015, p.~6.

\bibitem{taigman2014deepface}
Y.~Taigman, M.~Yang, M.~Ranzato, and L.~Wolf, ``Deepface: Closing the gap to
  human-level performance in face verification,'' in \emph{Proceedings of the
  IEEE conference on computer vision and pattern recognition}, 2014, pp.
  1701--1708.

\bibitem{schroff2015facenet}
F.~Schroff, D.~Kalenichenko, and J.~Philbin, ``Facenet: A unified embedding for
  face recognition and clustering,'' in \emph{Proceedings of the IEEE
  Conference on Computer Vision and Pattern Recognition}, 2015, pp. 815--823.

\bibitem{wright2009robust}
J.~Wright, A.~Y. Yang, A.~Ganesh, S.~S. Sastry, and Y.~Ma, ``Robust face
  recognition via sparse representation,'' \emph{IEEE transactions on pattern
  analysis and machine intelligence}, vol.~31, no.~2, pp. 210--227, 2009.

\bibitem{baltrusaitis2013constrained}
T.~Baltrusaitis, P.~Robinson, and L.-P. Morency, ``Constrained local neural
  fields for robust facial landmark detection in the wild,'' in
  \emph{Proceedings of the IEEE International Conference on Computer Vision
  Workshops}, 2013, pp. 354--361.

\bibitem{cristinacce2006feature}
D.~Cristinacce and T.~F. Cootes, ``Feature detection and tracking with
  constrained local models.'' in \emph{BMVC}, vol.~1, no.~2, 2006, p.~3.

\bibitem{prince2012computer}
S.~J. Prince, \emph{Computer vision: models, learning, and inference}.\hskip
  1em plus 0.5em minus 0.4em\relax Cambridge University Press, 2012.

\bibitem{gartner2014computational}
B.~G{\"a}rtner and M.~Hoffmann, ``Computational geometry -- lecture notes hs
  2013, chapter 6,'' ETH Z{\"u}rich University, Tech. Rep., 2014.

\bibitem{webYaleDataset}
``Yale face dataset,'' \url{http://vision.ucsd.edu/content/yale-face-database},
  accessed: 2017-17-1.

\bibitem{hastie2002elements}
T.~Hastie, R.~Tibshirani, and J.~Friedman, ``The elements of statistical
  learning: Data mining, inference, and prediction,'' \emph{Biometrics}, 2002.

\bibitem{bishop2007pattern}
C.~Bishop, ``Pattern recognition and machine learning (information science and
  statistics), 1st edn. 2006. corr. 2nd printing edn,'' \emph{Springer, New
  York}, 2007.

\bibitem{perlibakas2004distance}
V.~Perlibakas, ``Distance measures for pca-based face recognition,''
  \emph{Pattern Recognition Letters}, vol.~25, no.~6, pp. 711--724, 2004.

\bibitem{mohammadzade2013projection}
H.~Mohammadzade and D.~Hatzinakos, ``Projection into expression subspaces for
  face recognition from single sample per person,'' \emph{IEEE Transactions on
  Affective Computing}, vol.~4, no.~1, pp. 69--82, 2013.

\bibitem{webAttDataset}
``At\&t face dataset,''
  \url{http://www.cl.cam.ac.uk/research/dtg/attarchive/facedatabase.html},
  accessed: 2017-17-1.

\bibitem{webCohnDataset}
``Cohn-kanade face dataset,'' \url{http://www.pitt.edu/~emotion/ck-spread.htm},
  accessed: 2017-17-1.

\bibitem{kanade2000comprehensive}
T.~Kanade, J.~F. Cohn, and Y.~Tian, ``Comprehensive database for facial
  expression analysis,'' in \emph{Automatic Face and Gesture Recognition, 2000.
  Proceedings. Fourth IEEE International Conference on}.\hskip 1em plus 0.5em
  minus 0.4em\relax IEEE, 2000, pp. 46--53.

\bibitem{webLfwDataset}
``Lfw face dataset,'' \url{http://vis-www.cs.umass.edu/lfw/}, accessed:
  2017-17-1.

\bibitem{huang2007labeled}
G.~B. Huang, M.~Ramesh, T.~Berg, and E.~Learned-Miller, ``Labeled faces in the
  wild: A database for studying face recognition in unconstrained
  environments,'' Technical Report 07-49, University of Massachusetts, Amherst,
  Tech. Rep., 2007.

\end{thebibliography}

\end{document}